\documentclass[runningheads,a4paper]{llncs}

\usepackage{amssymb}
\usepackage{graphicx}
\usepackage{amsmath}
\usepackage{color}
\usepackage{subfigure}
\usepackage{multirow}
\usepackage{array}
\usepackage{microtype}
\usepackage{amsmath,bm}
\usepackage{epstopdf}
\usepackage{cite}
\usepackage{booktabs}
\usepackage{bbding}

\usepackage{url}

\makeatletter
\newcommand*{\textoverline}[1]{$\overline{\hbox{#1}}\m@th$}
\makeatother

\urldef{\mailsa}\path|{alfred.hofmann, ursula.barth, ingrid.haas, frank.holzwarth,|
\urldef{\mailsb}\path|anna.kramer, leonie.kunz, christine.reiss, nicole.sator,|
\urldef{\mailsc}\path|erika.siebert-cole, peter.strasser, lncs}@springer.com|

\begin{document}

\mainmatter  

\title{Semi-Automatic RECIST Labeling on CT Scans with Cascaded Convolutional Neural Networks}

\titlerunning{Semi-Automatic RECIST Labeling on CT scans with Cascaded CNNs}

%
%
\author{Youbao Tang\inst{1}, Adam P. Harrison\inst{2}, Mohammadhadi Bagheri\inst{3}, Jing Xiao\inst{4}, Ronald M. Summers\inst{1}}%
\authorrunning{Youbao Tang, \textit{et al}.}

\institute{$^1$ Imaging Biomarkers and Computer-Aided Diagnosis Laboratory\\
\email{youbao.tang@nih.gov}\\
$^3$ Clinical Image Processing Service \\
National Institutes of Health Clinical Center, Bethesda, MD 20892, USA \\
$^2$ NVIDIA, Santa Clara, CA 95051, USA \\
$^4$ Ping An Insurance Company of China, Shenzhen, 510852, China\\
}

%
%

\maketitle

\begin{abstract}
Response evaluation criteria in solid tumors (RECIST) is the standard measurement for tumor extent to evaluate treatment responses in cancer patients. As such, RECIST annotations must be accurate. However, RECIST annotations manually labeled by radiologists require professional knowledge and are time-consuming, subjective, and prone to inconsistency among different observers. To alleviate these problems, we propose a cascaded convolutional neural network based method to semi-automatically label RECIST annotations and drastically reduce annotation time. The proposed method consists of two stages: lesion region normalization and RECIST estimation. We employ the spatial transformer network (STN) for lesion region normalization, where a localization network is designed to predict the lesion region and the transformation parameters with a multi-task learning strategy. For RECIST estimation, we adapt the stacked hourglass network (SHN), introducing a relationship constraint loss to improve the estimation precision. STN and SHN can both be learned in an end-to-end fashion. We train our system on the DeepLesion dataset, obtaining a consensus model trained on RECIST annotations performed by multiple radiologists over a multi-year period. Importantly, when judged against the inter-reader variability of two additional radiologist raters, our system performs more stably and with less variability, suggesting that RECIST annotations can be reliably obtained with reduced labor and time.
\end{abstract}

\section{Introduction}
\begin{figure}[t!]
	\begin{center}
		\subfigure[]{\includegraphics[height=0.19\linewidth]{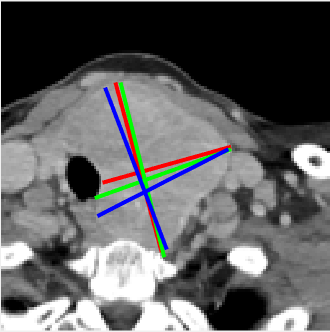}}
		\label{fig:annotation-a}
		\subfigure[]{\includegraphics[height=0.19\linewidth]{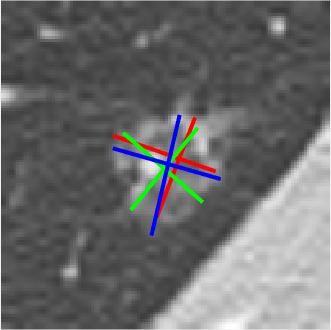}}
		\label{fig:annotation-b}
		\subfigure[]{\includegraphics[height=0.19\linewidth]{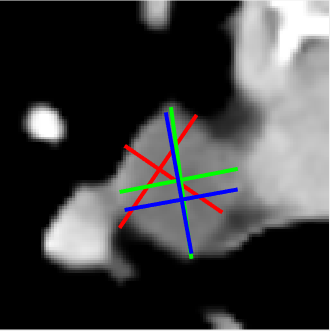}}
		\label{fig:annotation-c}
		\subfigure[]{\includegraphics[height=0.19\linewidth]{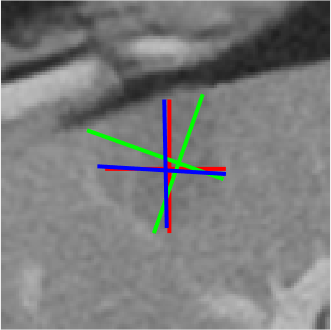}}
		\label{fig:annotation-d}
		\subfigure[]{\includegraphics[height=0.19\linewidth]{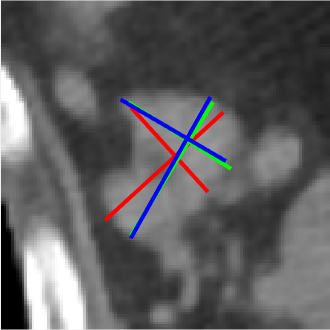}}
		\label{fig:annotation-e}
	\end{center}
	\caption{Five examples of RECIST annotations labeled by three radiologists. For each image, the RECIST annotations from different observers are indicated by diameters with different colors. Better viewed in color.}
	\label{fig:annotation}
\end{figure}

Response evaluation criteria in solid tumors (RECIST)~\cite{eisenhauer2009new} measures lesion or tumor growth rates across different time points after treatment. Today, the majority of clinical trials evaluating cancer treatments use RECIST as an objective response measurement \cite{kaisary2016lecture}. Therefore, the quality of RECIST annotations will directly affect the assessment result and therapeutic plan. Also, the RECIST annotation can be used as a weakly supervisory cue for lesion segmentation \cite{cai2018lesion}. To perform RECIST annotations, a radiologist first selects an axial image slice where the lesion has the longest spatial extent. Then he or she measures the diameters of the in-plane longest axis and the orthogonal short axis. These two axes constitute the RECIST annotation. Fig. \ref{fig:annotation} depicts five examples of RECIST annotations labeled by three different radiologists with different colors.

Using RECIST annotation face two main challenges. 1) measuring tumor diameters requires a great deal of professional knowledge and is time-consuming. Consequently, it is difficult and expensive to manually annotate large-scale datasets, e.g., those used in large clinical trials or retrospective analyses.
2) RECIST marks are often subjective and prone to inconsistency among different observers \cite{yoon2016observer}. For instance, from Fig. \ref{fig:annotation}, we can see that there is large variation between RECIST annotations from different radiologists. However, consistency is critical in assessing actual lesion growth rates, which directly impacts patient treatment options \cite{yoon2016observer}. To overcome these problems, we propose a RECIST estimation method that uses a cascaded convolutional neural network (CNN) approach. Given region of interest (ROI) cropped using a bounding box roughly drawn by a radiologist, the proposed method directly outputs RECIST annotations. As a result, the proposed RECIST estimation method is semi-automatic, drastically reducing annotation time while keeping the ``human in the loop''. To the best of our knowledge, this paper is the first to propose such an approach. In addition, our method can be readily made fully automatic as it can be trivially connected with any effective lesion localization framework. 

From Fig. \ref{fig:annotation}, the endpoints of RECIST annotations can well represent their locations and sizes. Thus, the proposed method estimates four keypoints, i.e., the endpoints, instead of two diameters. Recently, many approaches \cite{newell2016stacked,8100084,8099626,8237406} have been proposed to estimate the keypoints of the human body, e.g., knee, ankle, and elbow, which is similar to our task. Inspired by the success and simplicity of stacked hourglass networks (SHN)~\cite{newell2016stacked} for human pose estimation, this work employs SHN for RECIST estimation. Because the long and short diameters are orthogonal, a new relationship constraint loss is introduced to improve the accuracy of RECIST estimation. Regardless of class, the lesion regions may have large variability in sizes, locations and orientations in different images. To make our method robust to these variations, the lesion region first needs to be normalized before feeding into the SHN. In this work, we use the spatial transformer network (STN)~\cite{jaderberg2015spatial} for lesion region normalization, where a ResNet-50~\cite{he2016deep} based localization network is designed for lesion region and transformation parameter prediction. 
Experimental results over the DeepLesion dataset~\cite{yan2017deeplesion} compare our method to the multi-rater annotations in that dataset, plus annotations from two additional radiologists. Importantly, our method closely matches the multi-rater RECIST annotations and, when compared against the two additional readers, exhibits less variability than the inter-reader variability.

In summary, this paper makes the following main contributions: 1) We are the first to automatically generate RECIST marks in a roughly labeled lesion region. 2) STN and SHN are effectively integrated for RECIST estimation, and enhanced using multi-task learning and an orthogonal constraint loss, respectively. 3) Our method evaluated on a large-scale lesion dataset achieves lower variability than manual annotations by radiologists.
\section{Methodology}
\begin{figure}[t!]
	\begin{center}
		\includegraphics[width=0.99\linewidth]{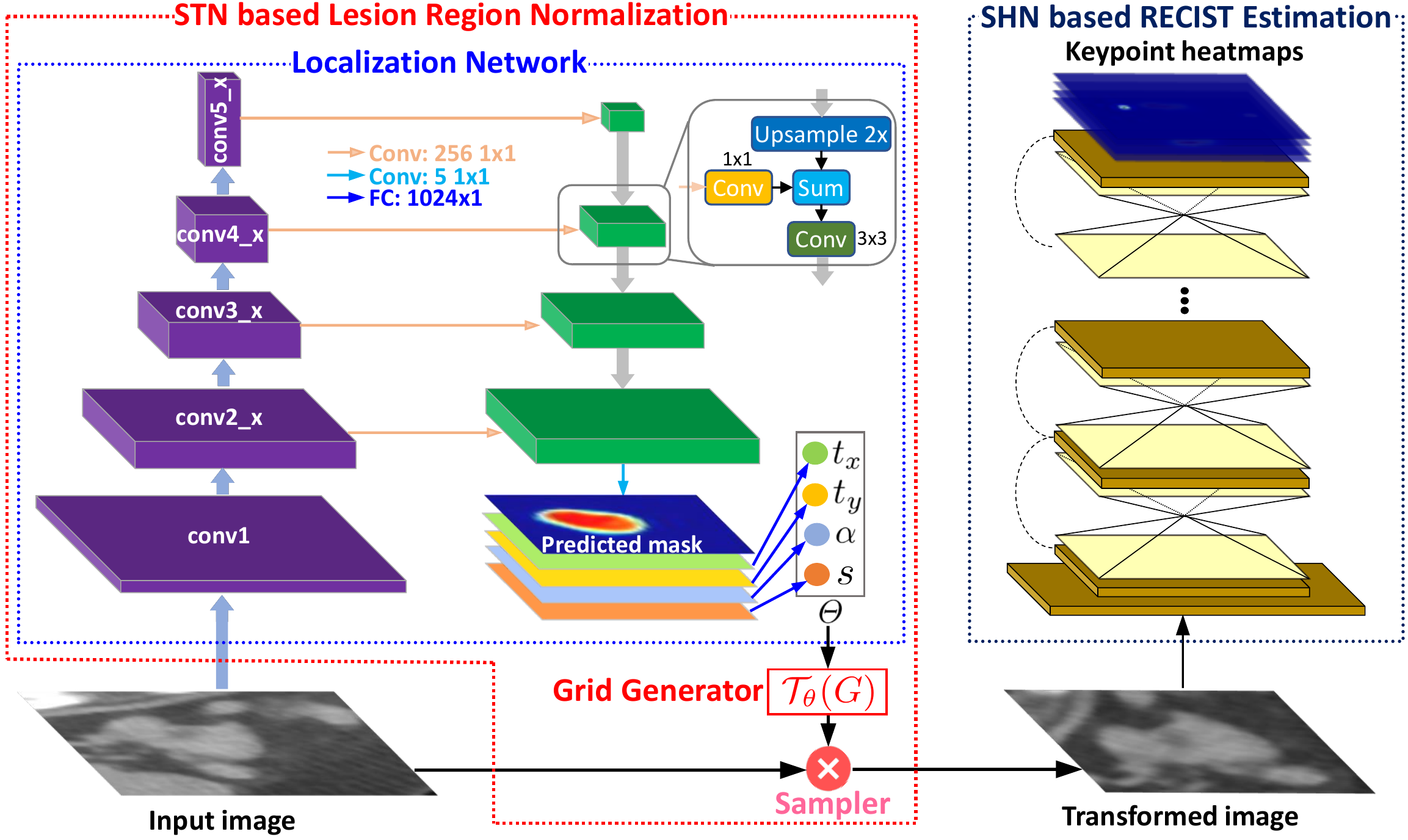}
	\end{center}
	\caption{The framework of the proposed method. The predicted mask and keypoint heatmaps are rendered with a color map for visualization purposes.}
	\label{fig:framework}
\end{figure}

Our system assumes the axial slice is already selected. To accurately estimate RECIST annotations, we propose a cascaded CNN based method, which consists of an STN for lesion region normalization and an SHN for RECIST estimation, as shown in Fig. \ref{fig:framework}. Here, we assume that every input image always contains a lesion region, which is roughly cropped by a radiologist. The proposed method can directly output an estimated RECIST annotation for every input.

\subsection{Lesion Region Normalization}
The original STN \cite{jaderberg2015spatial} contains three components, i.e., a localization network, a grid generator, and a sampler, as shown in Fig.~\ref{fig:framework}. The STN can implicitly predict transformation parameters of an image and can be used to implement any parameterizable transformation.
In this work, we use STN to explicitly predict translation, rotation and scaling transformations of the lesion. Therefore, the transformation matrix $\textbf{M}$ can be formulated as:
\begin{equation}
\small
\begin{aligned}
\textbf{M}=
    \overbrace{\left[ \begin{array}{ccc}
	1 & \ 0 & \ t_x\\
	0 & \ 1 & \ t_y\\
	0 & \ 0 & \ 1
	\end{array} 
	\right ]}^{Translation}
    \overbrace{\left[ \begin{array}{ccc}
	\cos(\alpha) & \ -\sin(\alpha) & \ 0\\
	\sin(\alpha) & \ \cos(\alpha) & \ 0\\
	0 & \ 0 & \ 1
	\end{array} 
	\right ]}^{Rotation}
    \overbrace{\left[ \begin{array}{ccc}
	s & \ 0 & \ 0\\
	0 & \ s & \ 0\\
	0 & \ 0 & \ 1
	\end{array} 
	\right ]}^{Scaling}=\left[ \begin{array}{ccc}
    s\cos(\alpha) & \ -s\sin(\alpha) & \ t_x\\
    s\sin(\alpha) & \ s\cos(\alpha) & \ t_y\\
    0 & \ 0 & \ 1
    \end{array} 
    \right ]
\end{aligned}
\label{equation:1}
\end{equation}

From \eqref{equation:1} there are four transformation parameters in $\textbf{M}$, denoted as $\theta=\{t_x,t_y,\alpha,s\}$. The goal of the localization network is to predict the transformation that will be applied to the input image. In this work, a localization network based on ResNet-50 \cite{he2016deep} is designed as shown in Fig.~\ref{fig:framework}. The purple blocks of Fig. \ref{fig:framework} are the first five blocks of ResNet-50. Importantly, unlike many applications of STN, the true $\theta$ can be obtained easily for transformation parameters prediction (TPP) by settling on a canonical layout for RECIST marks. 

As Sec.~\ref{sec:results} will outline, the STN also benefits from additional supervisory data, in the form of lesion pseudo-masks. To this end, we generate a lesion pseudo-mask by constructing an ellipse from the RECIST annotations. Ellipses are a rough analogue to a lesion's true shape. We denote this task lesion region prediction (LRP). Finally, to further improve prediction accuracy, we introduce another branch (green in Fig.~\ref{fig:framework}) to build a feature pyramid, similar to previous work~\cite{lin2017feature}, using a top-down pathway and skip connections.  The top-down feature maps are constructed using a ResNet-50-like structure. Coarse-to-fine feature maps are first upsampled by a factor of 2, and corresponding fine-to-coarse maps are transformed by 256 $1\times1$ convolutional kernels. These are summed, and resulting feature map will be smoothed using 256 $3\times3$ convolutional kernels. This ultimately produces a 5-channel $32\times32$ feature map, with one channel dedicated to the LRP. The remaining TPP channels are inputted to a fully connected layer outputting four transformation values, as shown in Fig. \ref{fig:framework}. 

According to the predicted $\theta$, a $2\times3$ matrix $\Theta$ can be calculated as
\begin{equation}
\Theta=
\left[ \begin{array}{ccc}
s\cos(\alpha) & \ -s\sin(\alpha) & \ t_x\\
s\sin(\alpha) & \ s\cos(\alpha) & \ t_y
\end{array} 
\right ]
\end{equation}
With $\Theta$, the grid generator $\mathcal{T}_\theta(G)$ will produce a parametrized sampling grid (PSG), which is a set of coordinates ${(x_i^s,y_i^s)}$ of source points where the input image should be sampled to get the coordinates ${(x_i^t,y_i^t)}$ of target points of the desired transformed image. Thus, the elements in PSG can be formulated as
\begin{equation}
\left[ \begin{array}{c}
    x_i^s\\
    y_i^s
    \end{array}\right ]=
    \left[ \begin{array}{ccc}
    s\cos(\alpha) & \ -s\sin(\alpha) & \ t_x\\
    s\sin(\alpha) & \ s\cos(\alpha) & \ t_y
    \end{array} 
    \right ]
    \left[ \begin{array}{c}
    x_i^t\\
    y_i^t\\
    1
    \end{array} 
    \right ]
\end{equation}

Armed with the input image and PSG, we use bilinear interpolation as a differentiable sampler to generate the transformed image. We set our canonical space to 1) center the lesion region, 2) make the long diameter  horizontal, and 3) remove most of THE background.
\subsection{RECIST Estimation}
After obtaining the transformed image, we need to estimate the positions of keypoints, i.e., the endpoints of long/short diameters. If the keypoints can be estimated precisely, RECIST annotation will be accurate. To achieve this goal, a network should have a coherent understanding of the whole lesion region and output high-resolution pixel-wise predictions. We use SHN~\cite{newell2016stacked} for this task, as they have the capacity to capture the above features and have been successfully used in human pose estimation. 

SHN is composed of stacked hourglass networks, where each hourglass network contains a downsampling and upsampling path, implemented by convolutional, max pooling, and upsampling layers. The topology of these two parts is symmetric, which means that for every layer present on the way down there is a corresponding layer going up and they are combined with skip connections. Multiple hourglass networks are stacked to form the final SHN by feeding the output of one as input into the next, as shown in Fig.~\ref{fig:framework}. Intermediate supervision is used in SHN by applying a loss at the heatmaps produced by each hourglass network, with the goal or improving predictions after each hourglass network. The outputs of the last hourglass network are accepted as the final predicted keypoint heatmaps. For SHN training, ground-truth keypoint heatmaps consist of four 2D Gaussian maps (with standard deviation of 1 pixel) centered on the endpoints of RECIST annotations. The final RECIST annotation is obtained according to the maximum of each heatmap. In addition, as the two RECIST axes should always be orthogonal, we also measure the cosine angle between them, which should always be 1. More details on SHN can found in Newell \textit{et al}.~\cite{newell2016stacked}. 
\subsection{Model Optimization}

We use mean squared error (MSE) loss to optimize our network, where all loss components are normalized into the interval $[0,1]$. The STN losses are denoted $L_{LRP}$ and $L_{TPP}$, which measure error in the predicted masks and transformation parameters, respectively. Training first focuses on LRP: $L_{STN}=10L_{LRP}+L_{TPP}$. After convergence, the loss focuses on the TPP: $L_{STN}=L_{LRP}+10L_{TPP}$. We first give a larger weight to $L_{LRP}$ to make STN focus more on LRP. After convergence, $L_{TPP}$ is weighted more heavily, so that the optimization is emphasized more on TPP. For SHN training, the losses are denoted $L_{HM}$ and $L_{cos}$, respectively, which measure error in the predicted heat maps and cosine angle, respectively. Each contribute equally to the total SHN loss. 

The STN and SHN networks are first trained separately and then combined for joint training. During joint training, all losses contribute equally. Compared with training jointly and directly from scratch, our strategy has faster convergence and better performance. We use stochastic gradient descent with a momentum of $0.9$, an initial learning rate of $5e^{-4}$, which is divided by $10$ once the validation loss is stable. After decreasing the learning rate twice, we stop training. To enhance robustness we augment data by random translations, rotations, and scales.

\begin{figure}[t!]
	\begin{center}
		\subfigure[]{\includegraphics[width=0.105\linewidth]{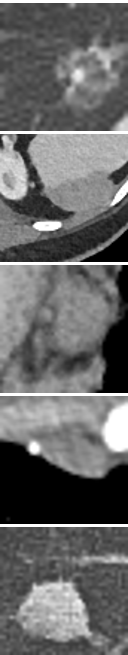}}
		\label{fig:result-a}
		\hspace{-0.013\linewidth}
		\subfigure[]{\includegraphics[width=0.105\linewidth]{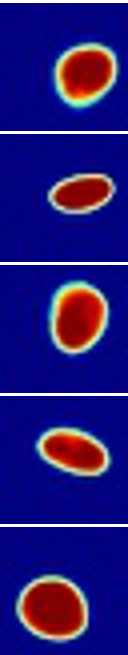}}
		\label{fig:result-b}
		\hspace{-0.013\linewidth}
		\subfigure[]{\includegraphics[width=0.105\linewidth]{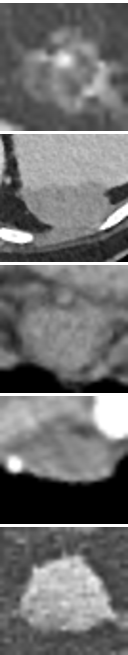}}
		\label{fig:result-c}
		\hspace{-0.013\linewidth}
		\subfigure[]{\includegraphics[width=0.105\linewidth]{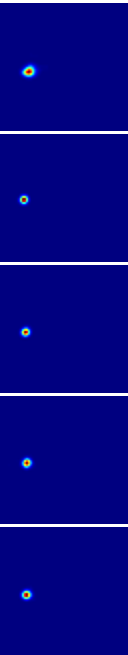}}
		\label{fig:result-d}
		\hspace{-0.013\linewidth}
		\subfigure[]{\includegraphics[width=0.105\linewidth]{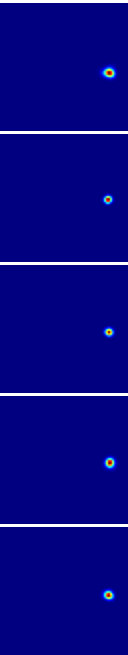}}
		\label{fig:result-e}
		\hspace{-0.013\linewidth}
		\subfigure[]{\includegraphics[width=0.105\linewidth]{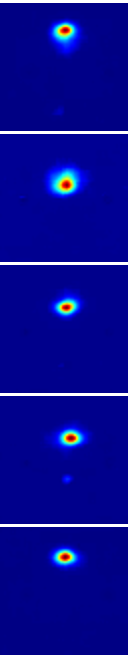}}
		\label{fig:result-f}
		\hspace{-0.013\linewidth}
		\subfigure[]{\includegraphics[width=0.105\linewidth]{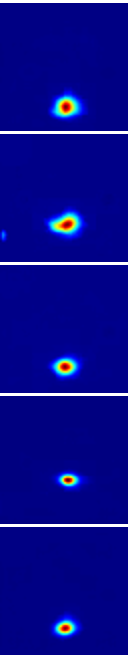}}
		\label{fig:result-g}
		\hspace{-0.013\linewidth}
		\subfigure[]{\includegraphics[width=0.105\linewidth]{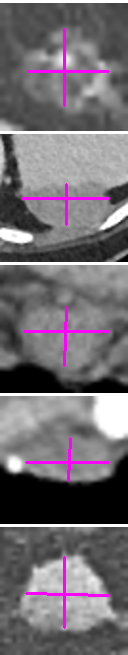}}
		\label{fig:result-h}
		\hspace{-0.013\linewidth}
		\subfigure[]{\includegraphics[width=0.105\linewidth]{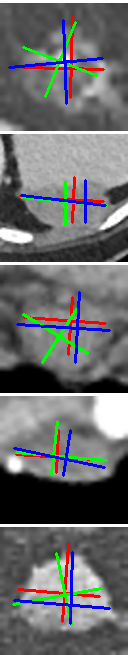}}
		\label{fig:result-i}
		\hspace{-0.013\linewidth}
	\end{center}
	\caption{Given the input test image (a), we can obtain the predicted lesion mask (b), the transformed image (c) from the STN, and the estimated keypoint heatmaps (d)-(g) from the SHN. From (d)-(g), we obtain the estimated RECIST (h), which is close to the annotations (i) labeled by radiologists. Red, green, and blue marks denote DL, R1, and R2 annotations, respectively.}
	\label{fig:result}
\end{figure}
\section{Experimental Results and Analyses}
\label{sec:results}
The proposed method is evaluated on the DeepLesion (DL) dataset~\cite{yan2017deeplesion}, which consists of $32,735$ images bookmarked and measured via RECIST annotations by multiple radiologists over multiple years from $10,594$ studies of $4,459$ patients. $500$ images are randomly selected from $200$ patients as a test set. For each test image, two extra RECIST annotations are labeled by another two experienced radiologists (R1 and R2). Images from the other $3,759$ and $500$ patients are used as training and validation datasets, respectively. To mimic the behavior of a radiologist  roughly drawing a bounding box around the entire lesion, input images are generated by randomly cropping a subimage whose region is $2$ to $2.5$ times as large as the lesion itself with random offsets. All images are resized to $128\times 128$. The performance is measured by the mean and standard deviation of the differences of keypoint locations and diameter lengths between RECIST estimations and radiologist annotations.

Fig. \ref{fig:result} shows five visual examples of the results. Fig. \ref{fig:result}(b) and \ref{fig:result}(c) demonstrate the effectiveness of our STN for lesion region normalization. With the transformed image (Fig. \ref{fig:result}(c)), the keypoint heatmaps (Fig. \ref{fig:result}(d)-(g)) are obtained using SHN. Figs.~\ref{fig:result}(d) and \ref{fig:result}(e) are the heatmaps of the left and right endpoints of long diameter, respectively, while Figs.~\ref{fig:result}(f) and \ref{fig:result}(g) are the top and bottom endpoints of the short diameter, respectively. Generally, the endpoints of long diameter can be found more easily than the ones of the short diameter, explaining why the highlighted spots in Figs.~\ref{fig:result}(d) and \ref{fig:result}(e) are smaller. As Fig.~\ref{fig:result}(h) demonstrates, the RECIST estimation correspond well with those of the radiologist annotations in Fig.~\ref{fig:result}(i). Note the high inter-reader variability.

\begin{table}[t!]
	\caption{The mean and standard deviation of the differences of keypoint locations (Loc.) and diameter lengths (Len.) between radiologist RECIST annotations and also those obtained by different experimental configurations of our method. The unit of all numbers is pixel in the original image resolution.}
	\linespread{1.4}
	\scriptsize
	\begin{center}
		\begin{tabular}{|@{}*{1}{m{1.75cm}<{\centering}@{}|@{}}*{8}{m{1.27cm}<{\centering}@{}|@{}}}
			\toprule
			\multirow{2}*{\textbf{Reader}}  & \multicolumn{2}{c|}{\textbf{DL} } & \multicolumn{2}{c|}{\textbf{R1}} & \multicolumn{2}{c|}{\textbf{R2}}  & \multicolumn{2}{c|}{ \textbf{Overall }}  \\ \cline{2-9}
			&  \textbf{Loc.} &  \textbf{Len.}  &  \textbf{Loc.}  &  \textbf{Len.}  & \textbf{Loc.}  &  \textbf{Len.}  &  \textbf{Loc.}  &  \textbf{Len.}  \\\toprule
			\multicolumn{9}{|c|}{\textbf{Long diameter}} \\\midrule
			 \textbf{DL}  &- &- &8.16$\pm$10.2 &4.11$\pm$5.87 &9.10$\pm$11.6 &5.21$\pm$7.42 & 8.63$\pm$10.9 & 4.66$\pm$6.71  \\ \hline 
			 \textbf{R1}  &8.16$\pm$10.2 &4.11$\pm$5.87 &- &- &6.63$\pm$11.0 &3.39$\pm$5.62 & 7.40$\pm$10.6 & 3.75$\pm$5.76  \\ \hline
			\textbf{R2}  &9.10$\pm$11.6 &5.21$\pm$7.42 &6.63$\pm$11.0 &3.39$\pm$5.62 &- &- & 7.87$\pm$11.3 & 4.30$\pm$6.65  \\ \hline
			 \textbf{\textoverline{SHN}}  &10.2$\pm$12.3 &6.73$\pm$9.42 &10.4$\pm$12.4 &6.94$\pm$9.83 &10.8$\pm$12.6 &7.13$\pm$10.4 & 10.5$\pm$12.5 & 6.93$\pm$9.87  \\ \hline
			 \textbf{\textoverline{STN}+\textoverline{SHN}} &7.02$\pm$9.43 &3.85$\pm$6.57 &7.14$\pm$11.4 &3.97$\pm$5.85 &8.74$\pm$11.2 &4.25$\pm$6.57 & 7.63$\pm$10.4 & 4.02$\pm$6.27  \\ \hline
			 \textbf{STN+\textoverline{SHN}}  &5.94$\pm$8.13 &3.54$\pm$5.18 &6.23$\pm$9.49 &3.62$\pm$5.31 &6.45$\pm$10.5 &3.90$\pm$6.21 & 6.21$\pm$9.32 & 3.69$\pm$5.59  \\ \hline
			 \textbf{STN+SHN}  &5.14$\pm$7.62 &3.11$\pm$4.22 &5.75$\pm$8.08 &3.27$\pm$4.89 &5.86$\pm$9.34 &3.61$\pm$5.72 & 5.58$\pm$8.25 & 3.33$\pm$4.93  \\ \midrule
			\multicolumn{9}{|c|}{\textbf{Short diameter}} \\\midrule
			\textbf{DL} &- &- &7.69$\pm$9.07 &3.41$\pm$4.72 &8.35$\pm$9.44 &3.55$\pm$5.24 & 8.02$\pm$9.26 & 3.48$\pm$4.99  \\ \hline 
			\textbf{R1} &7.69$\pm$9.07 &3.41$\pm$4.72 &- &- &6.13$\pm$8.68 &2.47$\pm$4.27 & 6.91$\pm$8.91 & 2.94$\pm$4.53  \\ \hline 
			\textbf{R2} &8.35$\pm$9.44 &3.55$\pm$5.24 &6.13$\pm$8.68 &2.47$\pm$4.27 &- &- & 7.24$\pm$9.13 & 3.01$\pm$4.81  \\ \hline 
			\textbf{\textoverline{SHN}}  &9.31$\pm$11.8 &5.02$\pm$7.04 &9.59$\pm$12.0 &5.19$\pm$7.35 &9.83$\pm$12.1 &5.37$\pm$7.69 & 9.58$\pm$11.8 & 5.19$\pm$7.38  \\ \hline
			\textbf{\textoverline{STN}+\textoverline{SHN}} &6.59$\pm$8.46 &3.25$\pm$5.93 &7.63$\pm$8.99 &3.35$\pm$6.41 &8.16$\pm$9.18 &4.18$\pm$6.48 & 7.46$\pm$8.93 & 3.59$\pm$6.22  \\ \hline
			\textbf{STN+\textoverline{SHN}} &5.52$\pm$7.74 &2.79$\pm$4.57 &5.71$\pm$8.06 &2.87$\pm$4.62 &6.01$\pm$8.39 &2.96$\pm$5.09 & 5.75$\pm$8.01 & 2.87$\pm$4.73  \\ \hline
			\textbf{STN+SHN} &4.47$\pm$6.26 &2.68$\pm$4.31 &4.97$\pm$7.02 &2.76$\pm$4.52 &5.41$\pm$7.59 &2.92$\pm$4.98 & 4.95$\pm$6.95 & 2.79$\pm$4.57  \\ \bottomrule
		\end{tabular}
	\end{center}
	\label{table:results-long}
\end{table}

To quantify this inter-reader variability, and how our approach measures against it, we compare the DL, R1, R2 annotations and those of our method  against each other, computing the mean and standard deviation of differences between axis locations and lengths. From the first three rows of each portion of Table \ref{table:results-long}, the inter-reader variability of each set of annotations can be discerned. The visual results in Figs.~\ref{fig:result}(h) and \ref{fig:result}(i) suggest that  our method corresponds well to the radiologists' annotations. To verify this, we compute the mean and standard deviation of the differences between the RECIST marks of our proposed method (STN+SHN) against those of three sets of annotations, as listed in the last row  of each part of Table \ref{table:results-long}. From the results, the estimated RECIST marks obtain the least mean difference and standard deviation in both location and length, suggesting the proposed method produces more stable RECIST annotations than the radiologist readers on the DeepLesion dataset. Note that the estimated RECIST marks are closest to the multi-radiologist annotations from the DL dataset, most likely because these are the annotations used to train our system. As such, this also suggest our method is able to generate a model that aggregates training input from multiple radiologists and learns a common knowledge that is not overfitted to any one rater's tendencies.

To demonstrate the benefits of our enhancements to standard STN and SHN, including the multi-task losses, we conduct the following experimental comparisons: 1) using SHN with only loss $L_{HM}$ (\textoverline{SHN}), which can be considered as the baseline; 2) using only the $L_{TPP}$ and $L_{HM}$ loss for the STN and SHN, respectively (denoted \textoverline{STN}+\textoverline{SHN}); 3) using both the $L_{TPP}$ and $L_{LRP}$ losses for the STN, but only the $L_{HM}$ loss for the SHN (STN+\textoverline{SHN}); 4) the proposed method with all $L_{TPP}$, $L_{LRP}$, $L_{HM}$, and $L_{cos}$ losses (STN+SHN). These results are listed in the last four rows of each part in Table \ref{table:results-long}. From the results, we can see that 1) the proposed method (STN+SHN) achieves the best performance. 2) \textoverline{STN}+\textoverline{SHN} outperforms \textoverline{SHN}, meaning that when lesion regions are normalized, the keypoints of RECIST marks can be estimated more precisely. 3) STN+\textoverline{SHN} outperforms \textoverline{STN}+\textoverline{SHN}, meaning the localization network with multi-task learning can predict the transformation parameters more precisely than with only a single task TPP. 4) STN+SHN outperforms STN+\textoverline{SHN},  meaning the accuracy of keypoint heatmaps can be improved by introducing the cosine loss to measure axis orthogonality. All of the above results demonstrate the effectiveness of the proposed method for RECIST estimation and the implemented modifications to improve performance.
\section{Conclusions}
We propose a semi-automatic RECIST labeling method that uses a cascaded CNN, comprised of enhanced STN and SHN. To improve the accuracy of transformation parameters prediction, the STN is enhanced using multi-task learning and an additional coarse-to-fine pathway. Moreover, an orthogonal constraint loss is introduced for SHN training, improving results further. The experimental results over the DeepLesion dataset demonstrate that the proposed method is highly effective for RECIST estimation, producing annotations with less variability than those of two additional radiologist readers. The semi-automated approach only requires a rough bounding box drawn by a radiologist, drastically reducing annotation time. Moreover, if coupled with a reliable lesion localization framework, our approach can be made fully automatic. As such, the proposed method can potentially provide a highly positive impact to clinical workflows.
\vspace*{0.5\baselineskip}

\noindent\textbf{Acknowledgments.}
This research was supported by the Intramural Research Program of the National Institutes of Health Clinical Center and by the Ping An Insurance Company through a Cooperative Research and Development Agreement. We thank Nvidia for GPU card donation.

%

\end{document}